\documentclass[conference]{IEEEtran}
\IEEEoverridecommandlockouts

\usepackage{cite}
\usepackage{amsmath,amssymb,amsfonts}
\usepackage{algorithmic}
\usepackage{graphicx}
\usepackage{textcomp}
\usepackage{xcolor}
\usepackage[a4paper, total={184mm,239mm}]{geometry}
\usepackage{mathtools}
\usepackage{systeme}
\usepackage[hyphens]{url}
\usepackage[hidelinks]{hyperref}
\usepackage{tikz}
\usetikzlibrary{shapes.misc, shapes.geometric, positioning, arrows.meta, chains}
\usepackage{adjustbox}
\usepackage[utf8]{inputenc}
\usepackage{subcaption}
\usepackage{lipsum}
\usepackage{balance}
\usepackage{float}
\usepackage{tabularx}

\begin{document}
\bstctlcite{MyBSTcontrol}
\title{BOMP-NAS: Bayesian Optimization Mixed Precision NAS}

\makeatletter
\newcommand{\newlineauthors}{%
  \end{@IEEEauthorhalign}\hfill\mbox{}\par
  \mbox{}\hfill\begin{@IEEEauthorhalign}
}
\makeatother

\author{
    \IEEEauthorblockN{David van Son}
    \IEEEauthorblockA{\textit{Eindhoven University of Technology}\\
                      Eindhoven, the Netherlands \\
                      d.v.son@tue.nl}
\and
    \IEEEauthorblockN{Floran de Putter}
    \IEEEauthorblockA{\textit{Eindhoven University of Technology}\\
                      Eindhoven, the Netherlands \\
                      f.a.m.d.putter@tue.nl}
\newlineauthors
    \IEEEauthorblockN{Sebastian Vogel}
    \IEEEauthorblockA{\textit{NXP Semiconductors}\\
                      Eindhoven, the Netherlands \\
                      sebastian.vogel@nxp.com}
\and
    \IEEEauthorblockN{Henk Corporaal}
    \IEEEauthorblockA{\textit{Eindhoven University of Technology}\\
                      Eindhoven, the Netherlands \\
                      h.corporaal@tue.nl}
}

\maketitle


\begin{abstract}
Bayesian Optimization Mixed-Precision Neural Architecture Search (BOMP-NAS) is an approach to quantization-aware neural architecture search (QA-NAS) that leverages both Bayesian optimization (BO) and mixed-precision quantization (MP) to efficiently search for compact, high performance deep neural networks. The results show that integrating quantization-aware fine-tuning (QAFT) into the NAS loop is a necessary step to find networks that perform well under low-precision quantization: integrating it allows a model size reduction of nearly 50\% on the CIFAR-10 dataset. BOMP-NAS is able to find neural networks that achieve state of the art performance at much lower design costs. This study shows that BOMP-NAS can find these neural networks at a 6$\times$ shorter search time compared to the closest related work.
\end{abstract}

\section{Introduction}

\label{sec:intro}
\IEEEPARstart{D}{eep learning} models have revolutionized image processing tasks, such as classification and semantic segmentation. However, designing these deep neural networks (DNNs) is a challenging task. It is especially challenging considering that nowadays, DNNs have to be deployed on resource constrained edge devices, e.g. mobile cellphones and electronic control units of cars. On these devices, DNNs are subject to limited memory and computational power constraints, while peak performance in terms of accuracy and latency is expected.

The proposed solution to the tedious task of network design is neural architecture search (NAS), an automated method of generating competitive DNN architectures. DNNs designed through NAS consistently outperform human\hbox{-}designed networks in various tasks both in terms of performance and efficiency.
 
Besides NAS, model compression techniques, such as architecture pruning and parameter quantization, have become an essential part of optimizing DNN architectures for embedded deployment \cite{coelho_automatic_2021, abs-1812-00090}. 
Using model compression, DNNs derived by NAS can be compressed even further, increasing efficiency while keeping performance intact.

This study introduces a sampling-based NAS methodology that integrates mixed-precision quantization and Bayesian optimization into a unified NAS algorithm, named Bayesian Optimization Mixed-Precision NAS (BOMP-NAS). Therefore, network quantization and the requirements and challenges for integrating quantization into the NAS optimization will be investigated in more detail in this paper. 

The contributions of this paper are:

\begin{enumerate}
\item A new sampling-based NAS methodology, called BOMP-NAS, with fine-grained mixed-precision (MP) quantization, where low-precision parameter use is enabled by quantization-aware fine-tuning (QAFT) during the search (Section \ref{sec:methods}).
\item Demonstrate the feasibility of BOMP-NAS as a quantization-aware NAS (QA-NAS) application, both in terms of found networks and search costs (Section \ref{sec:results}).
\item BOMP-NAS finds better performing models with similar memory budgets at 6$\times$ shorter search time compared to state-of-the-art (Section \ref{sec:discussion}).
\end{enumerate}

This paper is structured as follows: Section \ref{sec:related} summarizes existing approaches to QA-NAS, and the differences to BOMP-NAS. In Section \ref{sec:methods}, the methodology behind BOMP-NAS and experimental setup is described. The results obtained using BOMP-NAS are discussed in Section \ref{sec:results} and compared to existing works in Section \ref{sec:discussion}. Section \ref{sec:ablation} describes the ablation studies conducted. Finally, Section \ref{sec:conclusion} concludes this paper and gives possible directions for future research.

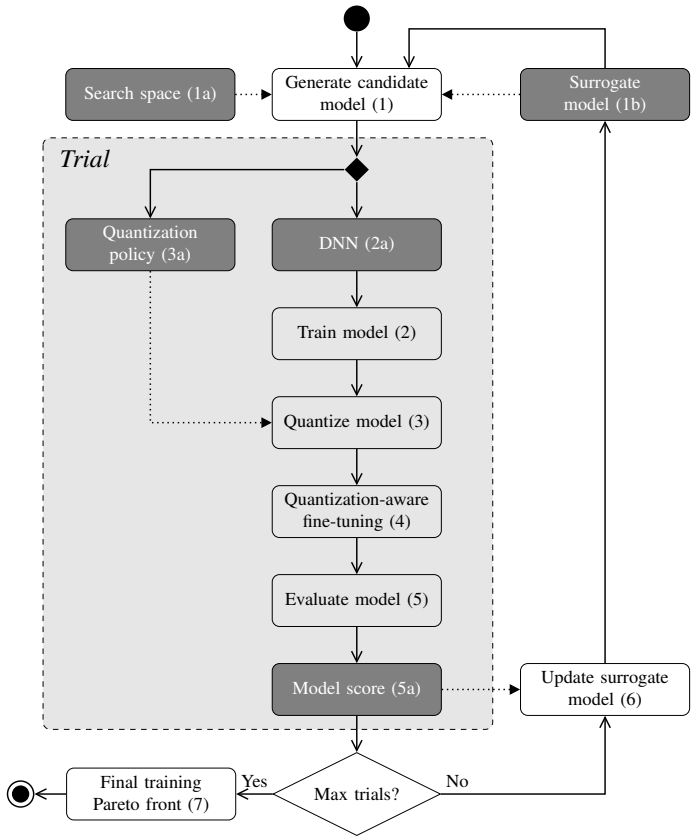
\begin{figure} 
\centering
\begin{adjustbox}{width=\linewidth}
\begin{tikzpicture}[  
  start/.style={circle, fill, minimum height=0.5cm},
  decision/.style={diamond, black, draw, text centered, text width=2.5cm, aspect=2},
  action/.style={rectangle, draw, rounded corners, text width=3cm, minimum height=1cm, text centered},
  var/.style={rectangle, draw, rounded corners, fill=black!50, text=white, text width=3cm, minimum height=1cm, text centered},
  arrow/.style={draw, -{Straight Barb[length=2mm, width=2mm]}, thick, label=above},
  dashedarrow/.style={draw, dotted, -{Triangle[length=2mm, width=2mm]}, thick},
  squarednode/.style={rounded rectangle, draw=black, very thick, text width=3cm, minimum height=1cm, align=center},
  fork/.style={diamond, fill},
  node distance=.7cm,
  ]
 
\draw [rounded corners, dashed, fill=black!10] (-6,-2.3) rectangle (2.6, -13.7); 

\node[]				(trial)	at (-5.2,-2.7)		{\textit{\Large Trial}};
\node[start]		(start)	[]				{};
\node[action]		(generate)	[below=of start]		{Generate candidate model (1)};
\node[var]			(space)	[left=of generate]			{Search space (1a)};
\node[fork]			(fork)		[below=of generate]		{};
\node[var]			(dnn)		[below=of fork]			{DNN (2a)};
\node[var]			(qpolicy)	[left=of dnn]			{Quantization policy (3a)};
\node[action]		(train)	[below=of dnn]				{Train model (2)};
\node[action]		(quantize)	[below=of train]		{Quantize model (3)};
\node[action]		(qaft)		[below=of quantize]		{Quantization-aware fine-tuning (4)};
\node[action]		(evaluate)	[below=of qaft]			{Evaluate model (5)};
\node[var]			(score)	[below=of evaluate]			{Model score (5a)};
\node[decision]		(trials_dec)	[below=of score, label=center:Max trials?]		{};
\node[action]		(update_bo)	[right=1.5cm of score]	{Update surrogate model (6)};
\node[var]			(strat)	[right=1.5cm of generate]	{Surrogate model (1b)};
\node[action]		(final)	[left=of trials_dec]		{Final training Pareto front (7)};
\node[circle, fill]	(end_small)	[left=of final]			{};
\node[] 			(temp) 	[right=of quantize] 		{};
\node[draw, circle, minimum height=.5cm, fill=none, thick]	(end)	at (end_small)	{};

\draw [arrow]	(start.south)		--	(generate.north);
\draw [arrow]	(generate.south)	--	(fork.north);
\draw [arrow]	(fork.south)		--	(dnn.north);
\draw [arrow]	(fork.west)		-|	(qpolicy.north);
\draw [arrow]	(dnn.south)		--	(train.north);
\draw [arrow]	(train.south)		--	(quantize.north);
\draw [arrow]	(quantize.south)	--	(qaft.north);
\draw [arrow]	(qaft.south)		--	(evaluate.north);
\draw [arrow]	(evaluate.south)	--	(score.north);
\draw [arrow]	(score.south)		--	(trials_dec.north);
\draw [arrow]	(trials_dec.east)	-|	node[pos=0.05, above, text centered] {No}	(update_bo.south);
\draw [arrow]	(update_bo.north)	--	(strat.south);
\draw [arrow]	(strat.north)		|-	(3, -0.2) -|	(1,-.96);
\draw [arrow]	(trials_dec.west)	--	node[midway, above, text centered] {Yes}	(final.east);
\draw [arrow]	(final.west)		--	(end.east);
\draw [dashedarrow]	(strat.west)		--	(generate.east);
\draw [dashedarrow]	(score.east)		--	(update_bo.west);
\draw [dashedarrow]	(space.east)		--	(generate.west);
\draw [dashedarrow]	(qpolicy.south)	|-	(quantize.west);
\end{tikzpicture}
\end{adjustbox}
\caption{UML activity diagram of proposed workflow of BOMP-NAS. DNNs (2a) and Quantization policies (3a) are selected (1) from the Search space (1a) using the Surrogate model (1b). The DNN is early trained in full precision (2), then quantized according to the (MP) Quantization policy. This quantized DNN is then fine-tuned quantization-aware (4). Next, the DNN is evaluated (5). These results are then scalarized into a score (5a) according to (\ref{eq:scorefn}). Lastly, the score is used to update the surrogate model (6), which is then used to sample the next candidate model (1).}
\label{fig:BOMP-NASworkflow}
\end{figure}

\section{Related work}
\label{sec:related}
Combining NAS with model compression techniques has proven an effective way to design compact DNNs that rival state-of-the-art (SotA) full precision networks. In several works, authors advocate for the joint optimization of DNN architecture and model compression \cite{abs-1811-09426, Wang2020CVPR, Kim2020, Bai2021}. This is because although the objectives can be pursued separately, this leads to sub optimal networks: i.e., the best architecture in a \texttt{float32} format floating point DNN may not be the best architecture in an \texttt{int8} format quantized DNN \cite{Wang2020CVPR}. 

In \cite{Kim2020}, a cell\hbox{-}based NAS was combined with homogeneous quantization-aware training (QAT) to generate compact, efficient DNNs. The authors first searched for efficient neural network building blocks, referred to as \textit{cells}, using gradient\hbox{-}based NAS, combined with gradient\hbox{-}based QAT. 

In \cite{Wang2020CVPR}, the once\hbox{-}for\hbox{-}all (OFA) \cite{abs-1908-09791} approach to NAS was used to quickly generate many different trained models, which could then be used to train their quantization\hbox{-}aware accuracy predictor. In this way, the search time is extremely short compared to other approaches, as the generated models do not need to be trained prior to evaluation. However, the initial investment of training the supergraph and accuracy predictor amounts to 2400 GPU hours on a V100 GPU, which is a significant investment barrier.

\cite{Bai2021} extends this work by introducing QAT into the training of the supergraph. The authors claim to have reduced the initial investment to 1805 hours. Compared to \cite{shen2020once}, the supernetwork required 300 epochs fewer training due to the introduction of BatchQuant, a method to reduce the instability of QAT when training supernetworks.

In \cite{Liberis2020}, aging evolution was combined with homogeneous PTQ to 8-bit to find networks suitable for microcontrollers. 
\cite{abs-1811-09426} extends this by also considering MP. In their work, an evolutionary algorithm\hbox{-}based NAS was combined with heterogeneous PTQ to 16, 8 or 4 bits. A significant limitation of this method is that the search engine is likely to get stuck in a bad local minimum. 

In all of these works, the results are compared against regular NAS networks, and quantized versions of existing architectures, which they consistently outperform. This shows that combining quantization and network architecture design into a single algorithm is a feasible approach for designing compact, high\hbox{-}performance networks.

The goal of this study is to reduce the time spent searching compared to existing quantization\hbox{-}aware NAS (QA\hbox{-}NAS) methods while integrating MP quantization to \{4-8\}\hbox{-}bit in sampling-based NAS. This study is most similar to \cite{abs-1811-09426}, with two major differences. Firstly, Bayesian optimization (BO) is used as the search strategy to traverse the search space more efficiently. This should reduce the search time of BOMP-NAS significantly compared to other methods, because BO converges quickly on promising solutions. Next to that, the issue of getting stuck on local minima is mitigated, since BO considers all previously trained networks, rather than a small subset as the population.

Secondly, BOMP-NAS uses QAFT, enabling DNNs to learn to compensate for the quantization noise. This should enable BOMP-NAS to derive DNNs that outperform SotA. 

This work is also somewhat similar to QFA \cite{Bai2021}, in that QFA and BOMP-NAS both use QAFT during NAS. However, BOMP-NAS is a sampling-based NAS approach, instead of the OFA-based approach of QFA, because of the investment barrier that is inherent in the OFA-based approaches.

\section{BOMP-NAS methodology} 
\label{sec:methods}

BOMP-NAS leverages multi-objective BO to efficiently search for MP DNNs. The proposed workflow of this approach is shown in Fig. \ref{fig:BOMP-NASworkflow}. The Search strategy (1b) selects (1) candidate DNNs (2a) and Quantization policies (3a) from the Search space (1a). 

The search space builds upon MobileNetV2 \cite{Sandler_2018_CVPR}, a compact, high-performing architecture originally designed for the ImageNet dataset. The MobileNetV2 architecture consists of a series of (possibly repeating) inverted bottlenecks. For each inverted bottleneck, the kernel size, width multiplier, expansion factor and number of repetitions was searchable. Also, for each layer within a bottleneck, the bitwidth is a searchable parameter. The search space is summarized in Table \ref{tab:searchspace}, and contains $3.96\cdot10^{19}$ architectures and $1.19\cdot10^{16}$ quantization policies. In total, the search space contains $4.73\cdot10^{39}$ MP DNNs.

With this search space, the aim is to find compact, high-performance networks for the CIFAR-10 and CIFAR-100 datasets \cite{Krizhevsky09learningmultiple}. The same search space was used for the CIFAR-100 dataset, except for the width multipliers, which could be chosen from [0.25, 0.50, \textbf{0.75}, 1.00, 1.30] instead. Since the CIFAR-10 and CIFAR-100 datasets are addressed, the image inputs are much smaller compared to the ImageNet dataset. Therefore, the resolution reduction occurs after bottlenecks 4 and 6 in the search space by means of a strided convolution, as proposed in \cite{elsken2019efficient}.

\begin{table}
\centering
\caption{Search space around MobileNetV2. The degrees of freedom are the kernel size \textit{\textbf{k}}, width multiplier \textit{\textbf{$\alpha$}}, expansion factor \textit{\textbf{e}} and the number of repetitions \textit{\textbf{n}}. The search space contains $3.96\cdot10^{19}$ architectures and $1.19\cdot10^{16}$ quantization policies. In total, the search space contains $4.73\cdot10^{39}$ MP DNNs. Choices indicated in bold are the seed values.}
\label{tab:searchspace}
\begin{tabular}{lll}
\textbf{Block}		&	\textbf{Parameter}		&	\textbf{Choices} \\ \hline
Inverted bottleneck 1	&	kernel size			&	[2, \textbf{3}, 4, 5, 6, 7]	\\
				&	width multiplier		&	[0.01, 0.05, \textbf{0.1}, 0.2, 0.3] \\
				&	expansion factor		&	[\textbf{1}] \\
				&	repetitions			&	[\textbf{1}] \\
Inverted bottleneck 2-6	&	kernel size			&	[2, \textbf{3}, 4, 5, 6, 7]	\\
				&	width multiplier		&	[0.01, 0.05, \textbf{0.1}, 0.2, 0.3] \\
				&	expansion factor		&	[1, 2, 3, 4, 5, \textbf{6}] \\
				&	repetitions			&	[0, \textbf{1}, 2, 3, 4, 5] \\
Inverted bottleneck 7	&	kernel size			&	[2, \textbf{3}, 4, 5, 6, 7]	\\
				&	width multiplier		&	[0.01, 0.05, \textbf{0.1}, 0.2, 0.3] \\
				&	expansion factor		&	[1, 2, 3, 4, 5, \textbf{6}] \\
				&	repetitions			&	[\textbf{1}] \\
Convolutional 2		&	number of filters		&	[128, 256, 512, 1024, \textbf{1280}]\\
				&	kernel size			&	[\textbf{1}]\\
				&	repetitions			&	[\textbf{1}]\\
Any				&	bitwidth			&	[4, 5, 6, 7, \textbf{8}]
\end{tabular}
\end{table}

For the search strategy, BOMP-NAS uses multi-objective Bayesian optimization (BO). because it exploits regularity in the search space very efficiently. Using only a few random initial datapoints, BO extracts the most promising candidate DNNs and Quantization policies, increasing the likelihood of finding good quantized DNNs in each trial. Following \cite{Jin2019}, BOMP-NAS uses a Gaussian process surrogate model (1b), with the Mat\'ern kernelization function to define edit-distances between DNN architectures. The acquisition function was chosen to be Upper Confidence Bound (UCB), again following \cite{Jin2019}.

The selected candidate DNN (2a) is trained in full precision. This performance estimation strategy, called early training, was also used in \cite{elsken2019efficient}. Early training provides a good relative ranking of each architecture at much lower cost than fully training each candidate DNN. Specifically, BOMP-NAS trains each candidate DNN (2a) for 20 epochs in full-precision (2). 

After the early training, the DNN is quantized (3) according to the Quantization policy (3a). Employing MP quantization in BOMP-NAS enables BOMP-NAS to distribute the available model size budget more carefully; important layers get higher precision, while less important layers get lower precision. The parameters of the DNNs are quantized per output channel, while activations were quantized per-tensor, as proposed in \cite{Nagel2021}.

The quantized DNN resulting from (3) is fine-tuned quantization-aware for 1 epoch (4). After the QAFT the early training is done, and the DNN can be evaluated (5a). The evaluation criteria in BOMP-NAS are flexible, and were chosen to be task accuracy [\%] and model size on disk [kB].

However, BO requires a single number to be returned as the score (5a) of a trial. Therefore, to enable multi-objectiveness, BOMP-NAS uses a scalarization function to combine multiple objectives into a single score. The notion of the scalarization function is to push for equal score along a Pareto front. This allows BOMP-NAS to show the trade-off between the optimization objectives that can be achieved for the current use-case. This is achieved by dividing the objectives into two categories: minimization and maximization objectives. 

For maximization objectives, the objective value, e.g. task accuracy, is divided by its corresponding reference value, e.g. \textit{ref\_accuracy}. For minimization objectives, the corresponding reference value, e.g. \textit{ref\_model\_size}, is divided by the reference value, e.g. disk size. In this way, a convex function is defined, which enables the generation of a Pareto front, instead of a single DNN as the result of NAS. The reference values can be tuned to increase or decrease importance of the objectives. The scalarization function BOMP-NAS uses is defined as:

\begin{equation}
\label{eq:scorefn}
score = \frac{accuracy\ [\%]}{ref\_accuracy} + \frac{ref\_model\_size}{log_{10}\left(model\ size\ [bits]\right)}
\end{equation}

The resulting score (5a) is used to update (6) the Surrogate model (1b), which is then used to sample the next candidate model (1). This cycle continues until the maximum number of trials has been reached. Lastly, the Pareto optimal DNNs are finally trained (7). During final training, the DNNs are trained for 200 epochs in full-precision, followed by 5 epochs of QAFT.

Within the BOMP-NAS workflow, it is possible to skip the QAFT during the search, this will be shown in the ablation studies (Section \ref{sec:ablation}). In the final training, also no QAFT is applied in this case. Both homogeneous and MP PTQ were investigated. Specifically, homogeneous (or fixed-precision) 8-bit quantization was compared against \{4,5,6,7,8\}-bit MP parameter quantization. For all experiments, the activations were quantized to \texttt{INT8}, and biases were quantized to \texttt{INT32}.



\subsection{Experimental setup}
\label{sec:setup}
BOMP-NAS combines the AutoKeras \cite{Jin2019} NAS framework with quantization provided by the QKeras \cite{coelho_automatic_2021} framework.

The baseline approach in this study is post-NAS quantization, also referred to as the NAS-then-quantize or sequential approach. In this approach, first a NAS is used to find the optimal full-precision architecture for a given problem; then, the optimal quantization policy for this network is determined. 


All searches were run for 100 iterations, from which only the Pareto optimal solutions in terms of task accuracy and model size were trained for 200 epochs to obtain the final Pareto front. The best models resulting from BOMP-NAS will be compared with quantized DNNs from related work on their task accuracy, disk size and design time. 

\section{Results}

\label{sec:results}
This section discusses the results obtained using BOMP-NAS. First, the Post-NAS PTQ baseline results are discussed, followed by the results of QAFT-aware NAS.

The baseline results were obtained by running BOMP-NAS on the search space defined in Table \ref{tab:searchspace} without quantization in the loop. After the NAS finished, all the networks were quantized homogeneously to 8-bit precision. 



The results of running BOMP-NAS with QAFT in the loop (Fig. \ref{fig:BOMP-NASworkflow}) on the search space are shown in Fig. \ref{fig:mp_qaft_cifar10}. 

The figure shows the model size [kB] (x-axis and blob size) and task accuracy [\%] (y-axis) of the candidate networks. The candidate networks are colored based on when they were sampled, earlier sampled models are darker than later sampled models. The networks sampled by BO should improve with time, as the surrogate model gets more information with each new sample. The finally trained Pareto optimal models are shown in red, with a line connecting them to their respective candidate network. The seed network shown is the one defined in Table \ref{tab:searchspace}. The dotted lines are equal-score lines, candidate networks along this line are considered equally optimal for the chosen reference values.

The figure shows that the found models are up to 2x smaller while achieving better accuracy than the seed architecture.

\begin{figure} 
\centering
\includegraphics[width=\linewidth]{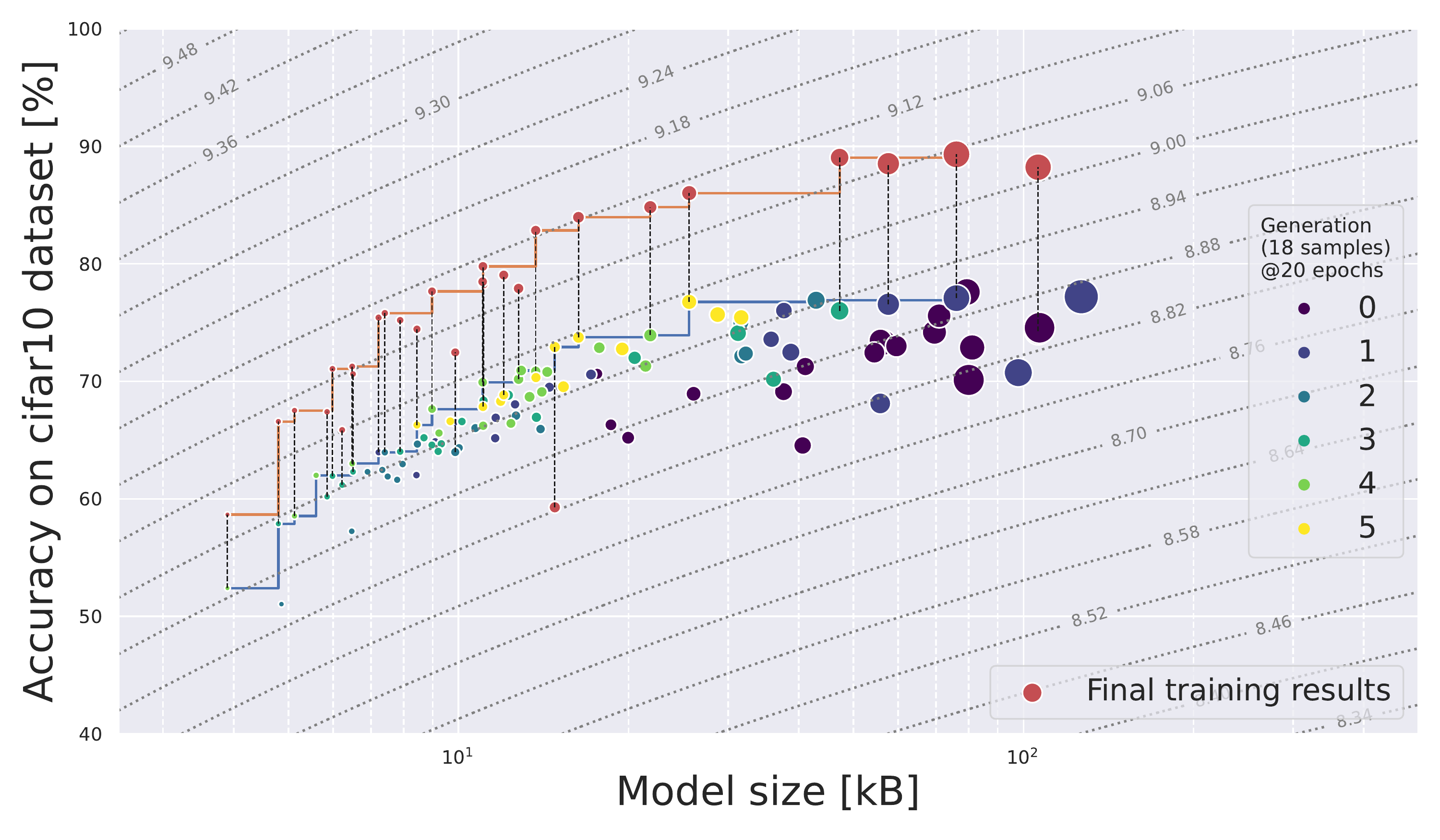}
\caption{Results of QAFT-aware NAS with $ref\_acc=0.8,\ ref\_model\_size=8$ on CIFAR-10. The found models are up to 2x smaller while achieving better accuracy than the seed architecture, which is MobileNetV2 quantized to 8-bit homogeneously.}
\label{fig:mp_qaft_cifar10}
\end{figure}

The bitwidth distributions per layer of the Pareto optimal models is shown in Fig. \ref{fig:bw_mp_qaft_cifar10}. It demonstrates that in this search, all of the models in the final Pareto front leverage the lower precision bitwidths available. This shows QAFT enables the leverage of low precision parameter quantization.

\begin{figure} 
\centering
\includegraphics[width=\linewidth]{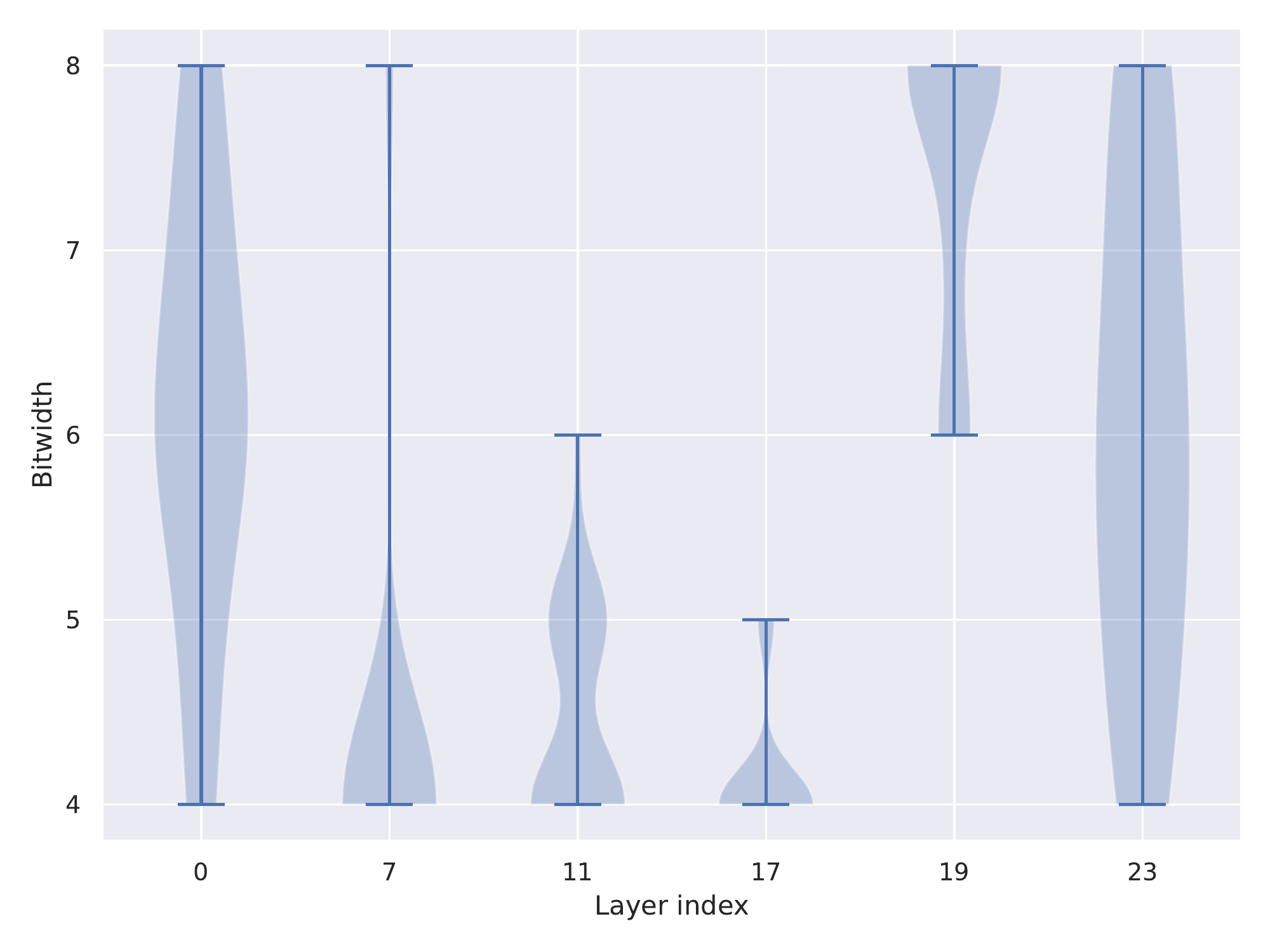}
\caption{Bitwidth distribution per layer for each of the models in the final Pareto front of the MP QAFT-aware NAS. The figure shows that in this search, all of the models in the final Pareto front leverage the lower precision bitwidths available. This shows QAFT enables the leverage of low precision parameter quantization.}
\label{fig:bw_mp_qaft_cifar10}
\end{figure}

The results of running BOMP-NAS on the CIFAR-100 search space are shown in Fig. \ref{fig:mp_qaft_cifar100}. 

\begin{figure} 
\centering
\includegraphics[width=\linewidth]{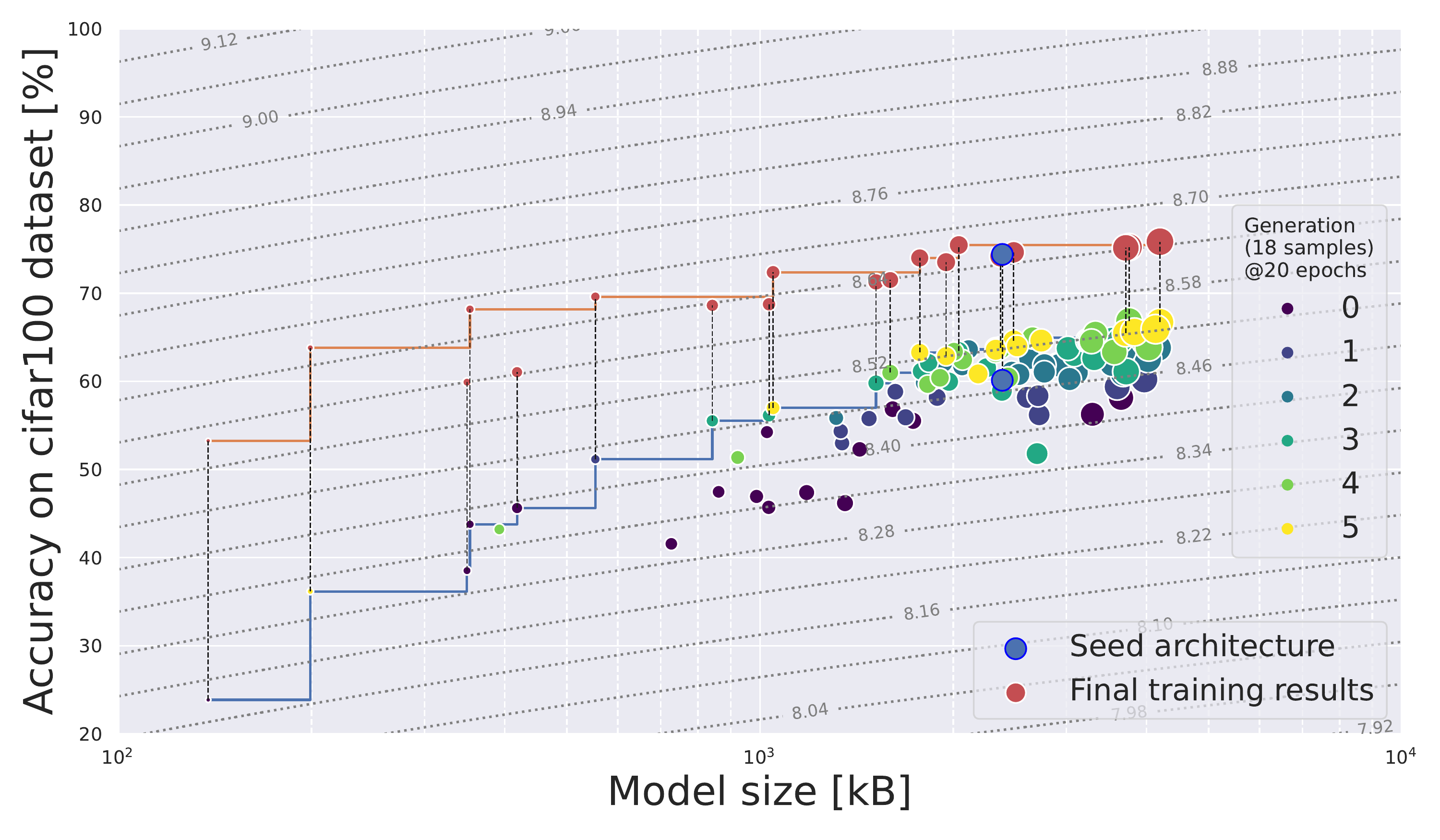}
\caption{Results of QAFT-aware NAS with $ref\_acc=0.8,\ ref\_model\_size=6$ on CIFAR-100.}
\label{fig:mp_qaft_cifar100}
\end{figure}

\section{Evaluation and Discussion}
\label{sec:discussion}
In this section, the results in Section \ref{sec:results} are compared with the baseline and works from SotA. First, the PTQ-aware NAS is compared to the baseline. Second, the effect of applying QAFT to networks found through PTQ-aware NAS is investigated. Lastly, the QAFT-aware NAS results are compared to the baseline and works from SotA in terms of performance and design cost.

Comparing the results from QAFT-NAS with the previously discussed Pareto fronts yields Fig. \ref{fig:mpptq_mpptqwithft_mpqat}. It shows that by integrating QAFT into the NAS, an even more optimal Pareto front can be obtained. BOMP-NAS now also finds many more promising models well below 10 kB disk size compared to the other approaches. 

\begin{figure} 
\centering
\includegraphics[width=\linewidth]{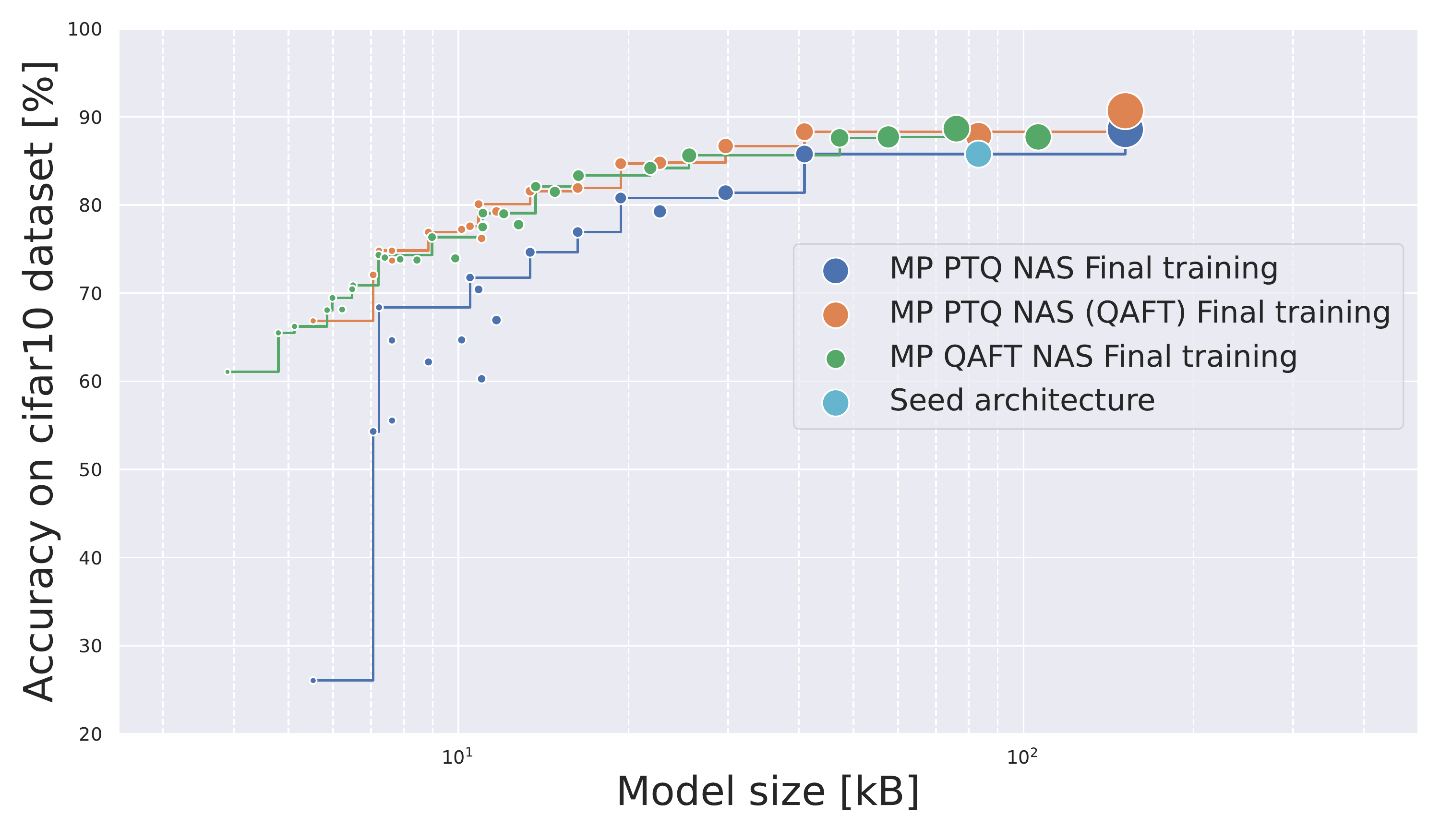}
\caption{Comparison between three Pareto fronts using {4-8}-bit MP quantization. The figure shows that fine-tuning the architectures found with PTQ search (MP PTQ-NAS (QAFT)) improves the results, especially on the left-hand side. However, QAFT-aware NAS yields even better results, especially on the left-hand side the performance of the found models is significantly improved.}
\label{fig:mpptq_mpptqwithft_mpqat}
\end{figure}

A comparison between the results of BOMP-NAS on CIFAR-10 and CIFAR-100, and state of the art is shown in Table \ref{tab:sota_comp}. The table shows that BOMP-NAS outperforms the reproduced version of JASQ on the same search space by more than 1pp with a similar model size. Compared to $\mu$NAS, BOMP-NAS performs 2.5pp worse, however, as shown in Table \ref{tab:searchcosts}, BOMP-NAS has a more than 40 times lower search time. 

For CIFAR-100, BOMP-NAS can outperform SotA works in the same size range in a single search. Due to the limited trials per search, it is expected that BOMP-NAS cannot outperform every baseline within a single search. The expectation is that, when considering models in the same size regime, BOMP-NAS can find better performing networks.

\begin{table} 
\centering
\caption{Pareto optimal architectures found by a single search of BOMP-NAS compared to works from SotA. The shown networks are the best performing networks that are smaller than or similar size as the respective SotA network. BOMP-NAS finds, in a single search, DNNs that outperform SotA in a broad model size range.}
\label{tab:sota_comp}
\begin{tabular}{clrr}
\textbf{Dataset}	&	\textbf{Method}			&	\textbf{Acc. [\%]}	&	\textbf{Model size [kB]} \\ \hline
CIFAR-10		&	JASQ	 (repr.)			&	65.97			&	4.47		\\
			&	BOMP-NAS					&	\textbf{67.36}	&	\textbf{4.57}		\\	\cline{2-4}
			&	JASQ \cite{abs-1811-09426}		&	97.03			&	900.00		\\
			&	BOMP-NAS					&	88.67			&	76.08		\\	\cline{2-4}
			&	$\mu$NAS \cite{Liberis2020}		&	\textbf{86.49}	&	\textbf{11.40}		\\
			&	BOMP-NAS					&	83.96			&	16.30		\\	\hline
CIFAR-100		&	DFQ \cite{Choi_2020_CVPR_Workshops}&	77.30			&	11200.00	\\
			&	GZSQ \cite{He_2021_CVPR}		&	75.95			&	5600.00	\\
			&	BOMP-NAS					&	75.84			&	4199.00		\\	\cline{2-4}
			&	LIE \cite{Liu_2021_CVPR}		&	73.34			&	1800.00		\\	
			&	BOMP-NAS					&	\textbf{74.00}	&	\textbf{1773.00}	\\	\cline{2-4}
			&	Mix\&Match \cite{Chang2020}		&	71.50			&	1700.00	\\
			&	LIE \cite{Liu_2021_CVPR}		&	71.24			&	1010.00	\\
			&	BOMP-NAS					&	\textbf{72.36}	&	\textbf{1047.00}	\\	\cline{2-4}
			&	APoT \cite{Li2019}			&	66.42			&	90.00		\\
			&	BOMP-NAS					&	68.18			&	353.00		
		
\end{tabular}
\end{table}

Table \ref{tab:searchcosts} shows a comparison between BOMP-NAS approach and SotA works. The table shows that, when compared to other QA-NAS methods on the same dataset, BOMP-NAS is significantly faster in yielding good results. An advantage of using BO is that, given a strict time budget, BOMP-NAS will converge faster on promising models compared to e.g. evolutionary approaches \cite{Jin2019}. Next to this, BOMP-NAS yields a Pareto front of trained DNNs, rather than a single network. This allows for better consideration of the trade-off between task accuracy and disk size.

\begin{table}
\centering
\caption{Search cost of various QA-NAS methods depending on the number of deployment scenarios N.}
\label{tab:searchcosts}
\begin{tabular}{c|cc}
\textbf{Method}			&	\textbf{Dataset}	&	\textbf{Search cost}	\\
					&				&	\textbf{(GPU hours)}	\\
\hline
APQ	\cite{Wang2020CVPR}		&	ImageNet		&	2400 + 0.5N	\\
OQA \cite{shen2020once}		&	ImageNet		&	1200 + 0.5N	\\
QFA \cite{Bai2021}			&	ImageNet		&	1805 + 0.N	\\
JASQ	 \cite{abs-1811-09426}	&	CIFAR10		&	72N		\\
$\mu$NAS \cite{Liberis2020}		&	CIFAR10		&	552N		\\
\hline
BOMP-NAS			&	CIFAR10		&	12N		\\
					&	CIFAR100		&	30N		\\
\end{tabular}
\end{table}

\section{Ablation studies}
\label{sec:ablation}
For the ablation studies, both MP PTQ-aware NAS and fixed-precision QAFT-aware NAS were evaluated, and compared to the results discussed in Section \ref{sec:results}.

Using the PTQ-aware NAS configuration of BOMP-NAS yields the results shown in Fig. \ref{fig:mp_ptq_cifar10}. The found networks on the far left-hand side perform significantly worse than the other models due to the extremely low bitwidths in these models. The search therefore focused on higher bitwidths, as is shown by the high model sizes of the found networks. This shows that simply applying MP PTQ to the found networks is not a good strategy to find efficient networks. Notable is that for the smallest model, applying PTQ after final training results in worse accuracy than applying PTQ after initial training. This shows that not only is the optimal quantization policy heavily dependent on the network architecture, also the current weight values have a significant influence. 

\begin{figure}
\centering
\includegraphics[width=\linewidth]{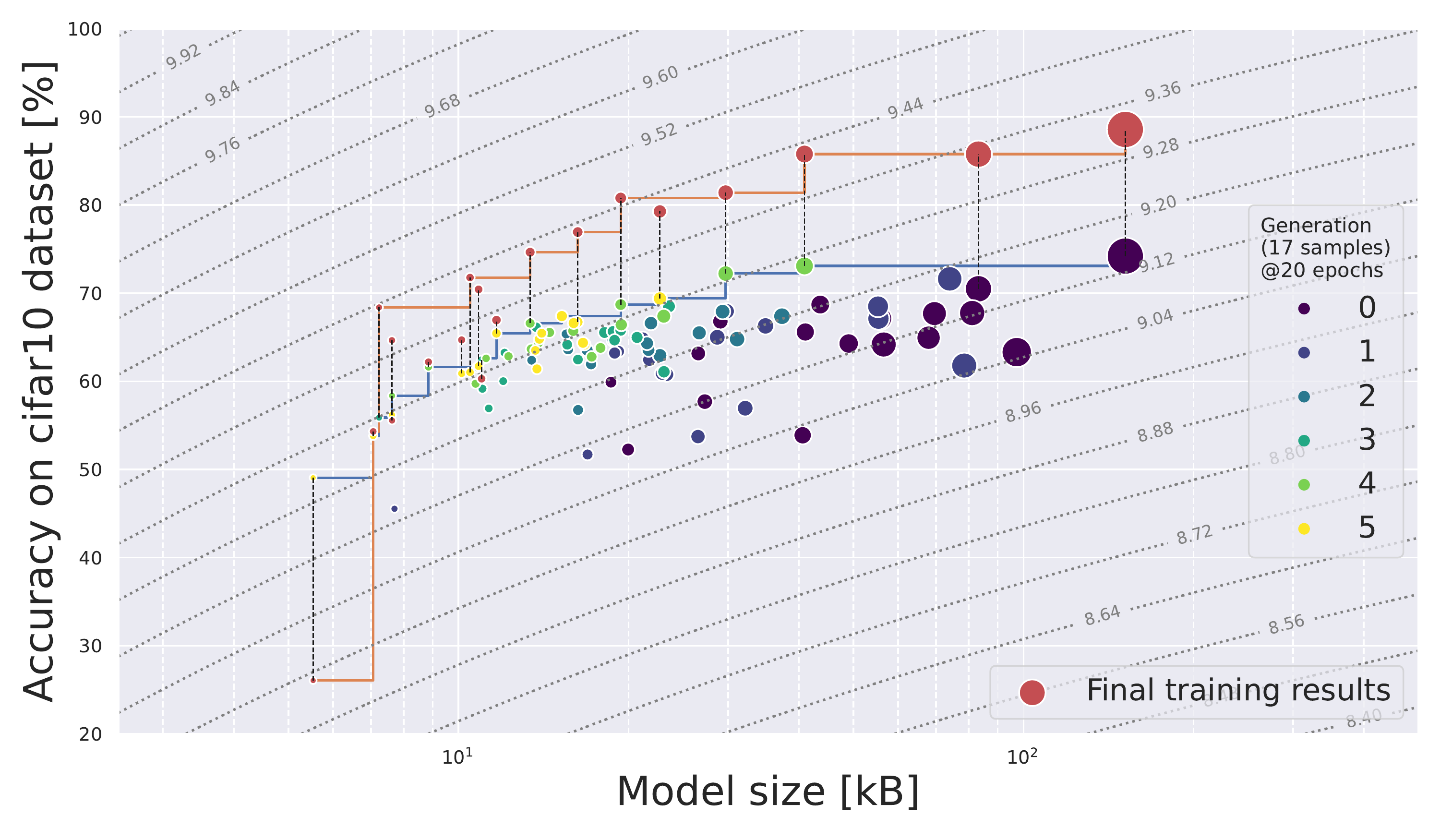}
\caption{Results of MP PTQ-aware NAS with $ref\_acc=0.8,\ ref\_model\_size=8$. The found networks on the far left-hand side perform significantly worse than the other models due to the extremely low bitwidths in these models. The search therefore focused on larger models, showing that simply applying MP PTQ to the found networks is not a good strategy to find efficient networks.}
\label{fig:mp_ptq_cifar10}
\end{figure}

A comparison between the results of 4-bit QAFT-NAS and the previously discussed Pareto fronts is shown in Fig. \ref{fig:paretofronts_all}, it shows that using fixed 4-bit quantization NAS was not always able to find better models compared to the MP approach. On the far left, the 4-bit approach can find more optimal networks, while in the middle of the size range, the networks generally perform worse than equally sized networks from other approaches. This could be due to the low sampling frequency that is observed in that range, as shown in Fig. \ref{fig:4b_qaft_cifar10}.

\begin{figure} 
\centering
\includegraphics[width=\linewidth]{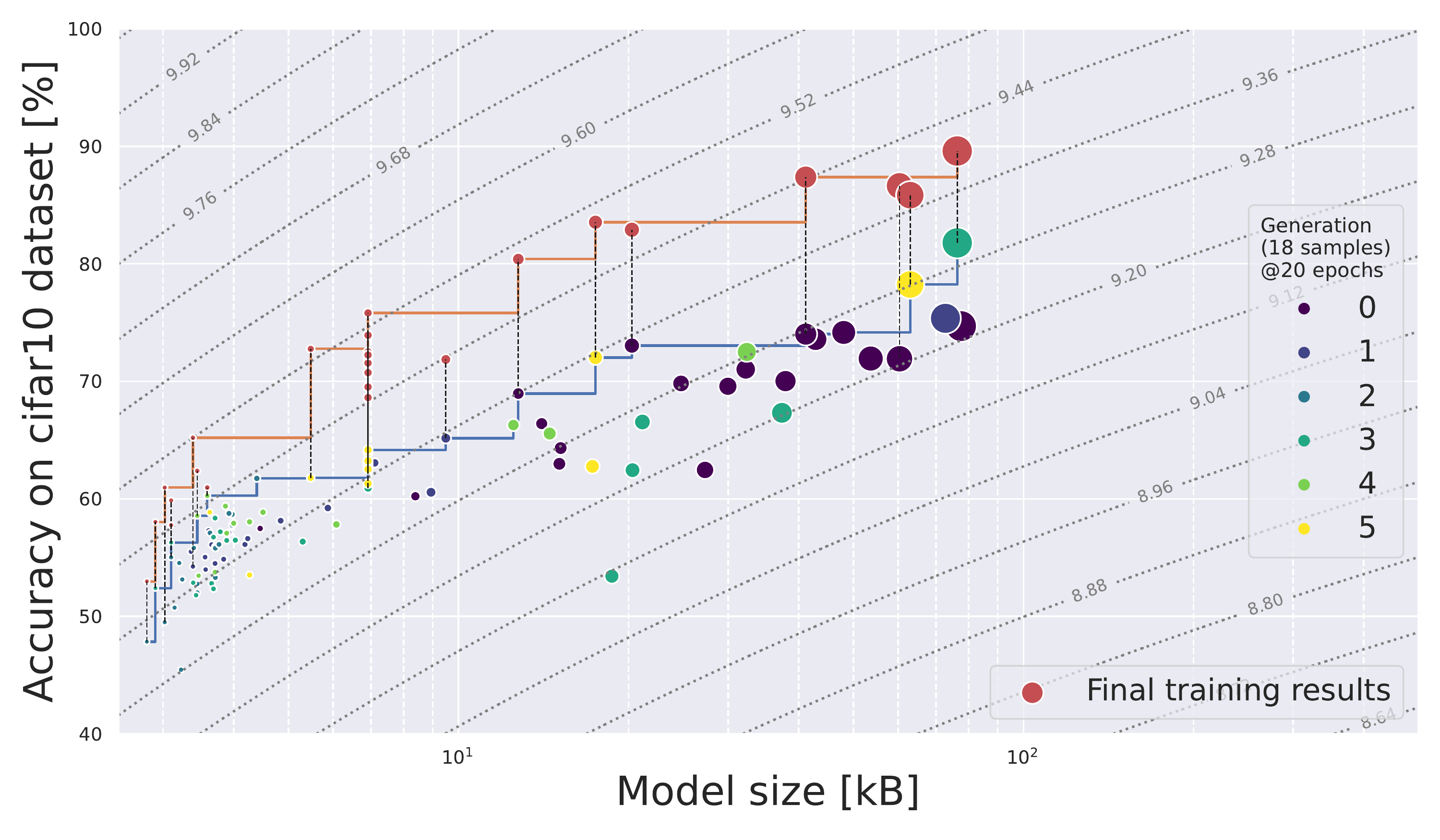}
\caption{Results of 4-bit QAFT-aware NAS with $ref\_acc=0.8,\ ref\_model\_size=8$.}
\label{fig:4b_qaft_cifar10}
\end{figure}
%

\begin{figure} 
\centering
\includegraphics[width=\linewidth]{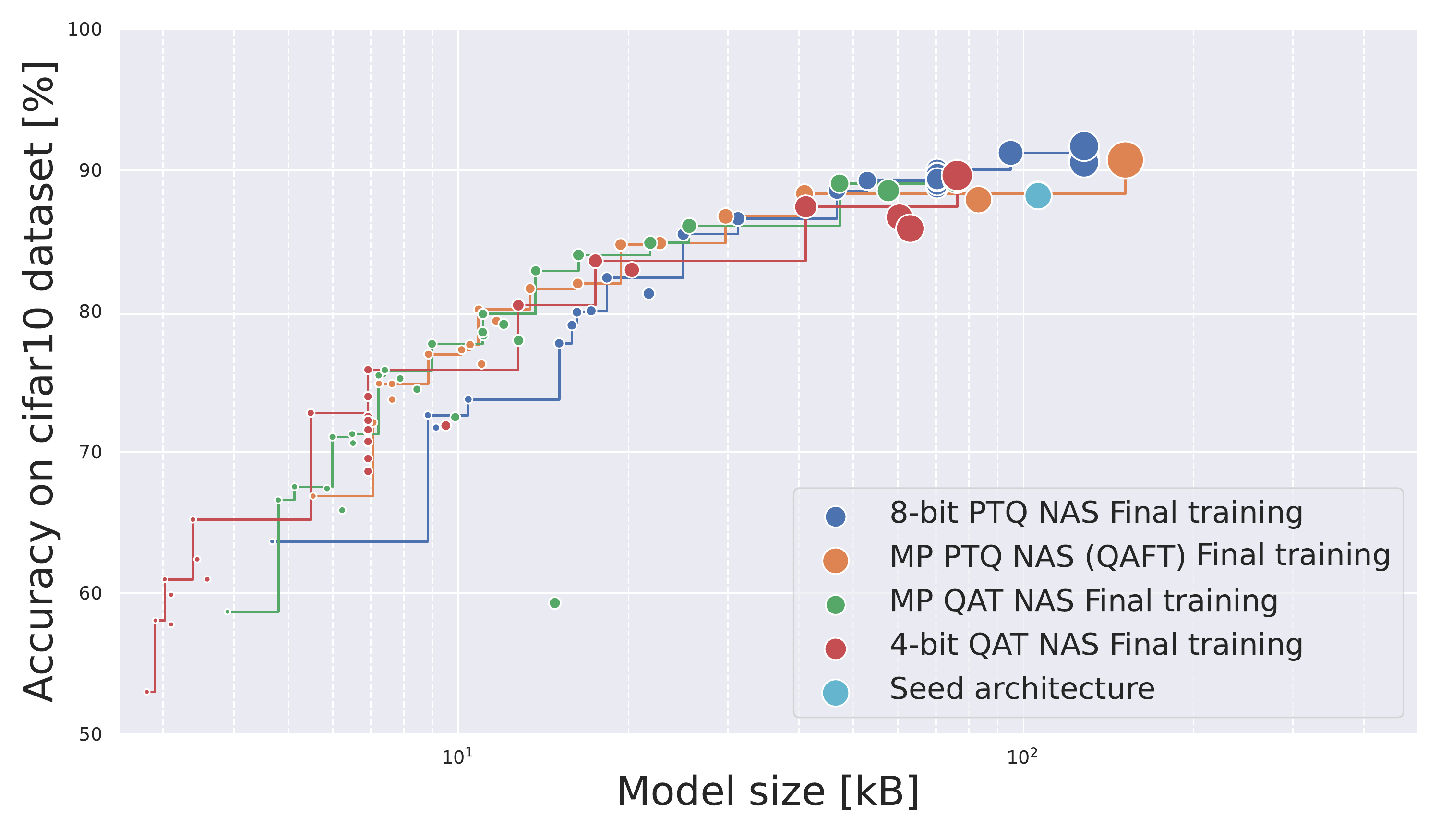}
\caption{Comparison between several Pareto fronts. The MP searches used bitwidths varying between 4 and 8-bit, fixed-bitwidth searches have the used bitwidth specified. The figure shows that using fixed 4-bit quantization NAS cannot always find better models compared to the MP approach, while PTQ-aware NAS even with post-NAS QAFT performs worse compared to QAFT-aware NAS.}
\label{fig:paretofronts_all}
\end{figure}

Table \ref{tab:searchcosts_bomp} shows a comparison between the search cost of the discussed QA-NAS approaches. The table shows that the introduction of MP into the search space does not affect the search time in BOMP-NAS. This is because the BO can heavily exploit its prior knowledge gained from previous samples due to the regularity inherent in quantization, therefore convergence time does not significantly increase. However, integrating QAFT into the NAS does significantly impact the search time. For example, when using MP QAFT-NAS, the search takes 25\% longer compared to the MP PTQ-aware NAS approach due to the added QAFT.

\begin{table}
\centering
\caption{Search cost of various QA-NAS approaches in BOMP-NAS depending on the number of deployment scenarios N.}
\label{tab:searchcosts_bomp}
\begin{tabular}{c|cc}
\textbf{Method}			&	\textbf{Dataset}	&	\textbf{Search cost}	\\
					&				&	\textbf{(GPU hours)}	\\
\hline
8-bit PTQ-aware NAS			&	CIFAR10		&	10N		\\
					&	CIFAR100		&	23N		\\
MP PTQ-aware NAS			&	CIFAR10		&	10N		\\
					&	CIFAR100		&	23N		\\
MP QAFT-aware NAS			&	CIFAR10		&	12N		\\
(BOMP-NAS)			&	CIFAR100		&	30N		\\
4-bit QAFT-aware NAS			&	CIFAR10		&	15N		\\
					&	CIFAR100		&	35N	
\end{tabular}
\end{table}

\section{Conclusion}
\label{sec:conclusion}
\balance
Designing deep neural networks is a challenging, but fundamental task in deep learning applications. To cater to the needs of edge devices, neural network design is often paired with model compression to design compact, high performance deep neural networks.  

Bayasian Optimization Mixed Precision (BOMP)-NAS is an approach to quantization-aware NAS that leverages both Bayesian optimization (BO) and mixed-precision (MP) to efficiently search for compact, high performance networks. BOMP-NAS is a an approach that allows the integration of quantization in NAS, and that can be integrated in NAS without much effort. 

This study shows that integrating QAFT into the NAS loop is a necessary step to find networks that perform well under low-precision quantization. Integrating QAFT in the loop allows BOMP-NAS to achieve a model size reduction of nearly 50\% on the CIFAR-10 dataset. Next to that, this paper shows that using BOMP-NAS, DNNs that achieve state of the art performance on the CIFAR datasets can be designed. For example, DNNs designed by BOMP-NAS outperform JASQ \cite{abs-1811-09426} by 1.4pp with a memory budget of 4.5 kB.

Lastly, this study shows that by using BO as the search strategy, BOMP-NAS finds state of the art models at much lower design costs. Compared to the closest related work, JASQ \cite{abs-1811-09426}, BOMP-NAS can find better performing models with similar memory budgets at 6$\times$ shorter search time.

For future research, a possible improvement could be to re-use the trained weights for each trial more efficiently. For example, for each trained full-precision network, multiple quantization policies could be tried. In this way, more information can be extracted from each trial, thereby reducing the search time further.

\bibliographystyle{IEEEtran}
\bibliography{References.bib}


\end{document}